\def\paperTitle{REDI-Match: Rotation-Equivariant Distillation for Efficient and Robust Dense Matching}

\def\authorBlock{%
    Yinji Ge$^{1}$\thanks{These authors contributed equally.}\qquad
    Guixu Zheng$^{2}$\footnotemark[1]\qquad
    Wulong Guo$^{3}$\qquad
    Qian Feng$^{4}$ \\[0.35cm]
    Xu Wu$^{1}$\qquad
    Kai Zhou$^{1}$\footnotemark[2]\qquad
    Xinyuan Liu$^{1}$\footnotemark[2]\qquad
    Fei Xing$^{1}$\thanks{Corresponding authors} \\[0.4cm]
    $^{1}$Tsinghua University \qquad $^{2}$Southern University of Science and Technology \\[0.1cm]
    $^{3}$Beihang University \qquad $^{4}$Zhejiang University \\[0.2cm]
    {\tt\small geyj25@mails.tsinghua.edu.cn, xingfei@mail.tsinghua.edu.cn}
}

\newif\ifreview 
\newif\ifarxiv \newcommand{\arxiv}{\arxivtrue}
\newif\ifcamera 
\newif\ifrebuttal 

\arxiv  % \review OR \arxiv OR \cameraready

\pdfoutput=1
\documentclass[10pt,twocolumn,letterpaper]{article}
\ifreview \usepackage[review]{cvpr} \fi
\ifarxiv \usepackage[pagenumbers]{cvpr} \fi
\ifrebuttal \usepackage[rebuttal]{cvpr} \fi
\ifcamera \usepackage{cvpr} \fi

\usepackage{graphicx}	
\usepackage{amsmath}	
\usepackage{amssymb}	
\usepackage{booktabs}
\usepackage{times}
\usepackage{microtype}
\usepackage{epsfig}
\usepackage{caption}
\usepackage{float}
\usepackage{placeins}
\usepackage{color, colortbl}
\IfFileExists{stfloats.sty}{\usepackage{stfloats}}{}
\usepackage{enumitem}
\usepackage{tabularx}
\usepackage{xstring}
\usepackage{multirow}
\usepackage{xspace}
\usepackage{url}
\usepackage{subcaption}
\usepackage[table]{xcolor}
\usepackage[hang,flushmargin]{footmisc}

\newcommand{\abltarget}[1]{\hypertarget{abl:#1}{\textcolor{cvprred}{\MakeUppercase{\romannumeral #1}}}}
\newcommand{\abllink}[1]{\hyperlink{abl:#1}{\textcolor{cvprred}{\MakeUppercase{\romannumeral #1}}}}

\ifcamera \usepackage[accsupp]{axessibility} \fi

% ...

\DeclareMathOperator*{\argmin}{arg\,min}

\newcommand{\R}[1]{{%
    \textbf{%
        \ifstrequal{#1}{1}{\textcolor{red}{R#1}}{%
        \ifstrequal{#1}{2}{\textcolor{blue}{R#1}}{%
        \ifstrequal{#1}{3}{\textcolor{magenta}{R#1}}{%
        \ifstrequal{#1}{4}{\textcolor{teal}{R#1}}{%
                           \textcolor{cyan}{R#1}%
        }}}}%
    }%
}}
\usepackage{xr-hyper}

\makeatletter
\newcommand*{\addFileDependency}[1]{
  \typeout{(#1)}
  \@addtofilelist{#1}
  \IfFileExists{#1}{}{\typeout{No file #1.}}
}

\makeatother

\definecolor{cvprblue}{rgb}{0.21,0.49,0.74}
\definecolor{cvprred}{rgb}{0.78,0.10,0.10}
\usepackage[pagebackref,breaklinks,colorlinks,citecolor=cvprblue,linkcolor=cvprred]{hyperref}
\usepackage[capitalize]{cleveref}
\crefname{section}{Sec.}{Secs.}
\crefname{table}{Table}{Tables}
\crefname{figure}{Fig.}{Figs.}

\ifarxiv \crefname{appendix}{App.}{Apps.}
\else \crefname{appendix}{Suppl.}{Suppls.} \fi

\usepackage{caption}

\begin{document}
%% TITLE
\title{\paperTitle}
\author{\authorBlock}
\maketitle

\begin{abstract}
% Abstract goes here.
Vision Foundation Models (VFMs) have significantly advanced dense feature matching, yet severe in-plane rotation remains a critical challenge. Existing solutions face a fundamental dilemma: data-driven methods require inefficient parameter scaling to implicitly learn rotations, whereas strictly equivariant networks lack the semantic capacity of modern VFMs. Consequently, current frameworks typically freeze VFMs and shift the entire burden of rotation generalization to the downstream decoder. To break this architectural bottleneck, we propose \textbf{REDI-Match}, an efficient framework driven by a novel Rotation-Equivariant Distillation (REDI) paradigm. Instead of relying on rotation data augmentation to establish rotational correspondences, REDI distills the non-equivariant semantic representations of a VFM into a lightweight, strictly rotation-equivariant encoder, leveraging an equivariant geometric architecture to constrain robust high-dimensional semantics. To fully exploit these features, we equip the decoder with an entropy-driven spatial alignment module. By evaluating discrete rotation hypotheses, this mechanism explicitly locks onto the canonical coordinate system, eliminating global ambiguity before continuous refinement. Extensive experiments demonstrate that REDI-Match establishes a new state-of-the-art (SOTA) across multiple benchmarks. Notably, it achieves a \textbf{13.89\%} absolute pose accuracy improvement on the highly challenging SatAst dataset while operating \textbf{1.9$\times$} faster than the current SOTA (RoMa v2), enabling real-time inference (\textbf{$\sim$41 FPS}) on a single RTX 4090 GPU. Code: \url{https://github.com/YinjiGe/REDI-Match}.

\end{abstract}
% Radar comparison chart (single column)
\begin{figure}[t]
    \centering
    \includegraphics[width=\columnwidth]{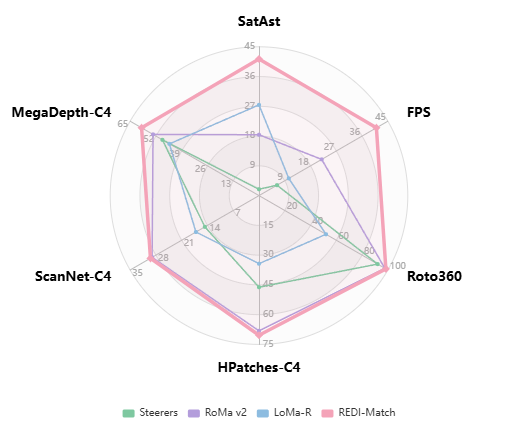}
    \caption{\textbf{Radar comparison of REDI-Match against SOTA methods.} All axes except FPS report AUC scores (higher is better). REDI-Match achieves the best overall accuracy–efficiency trade-off across all benchmarks.}
    \label{fig:radar}
\end{figure}

% SatAst qualitative comparison teaser (page 2 top)
\begin{figure*}[t]
    \centering
    \includegraphics[width=\textwidth]{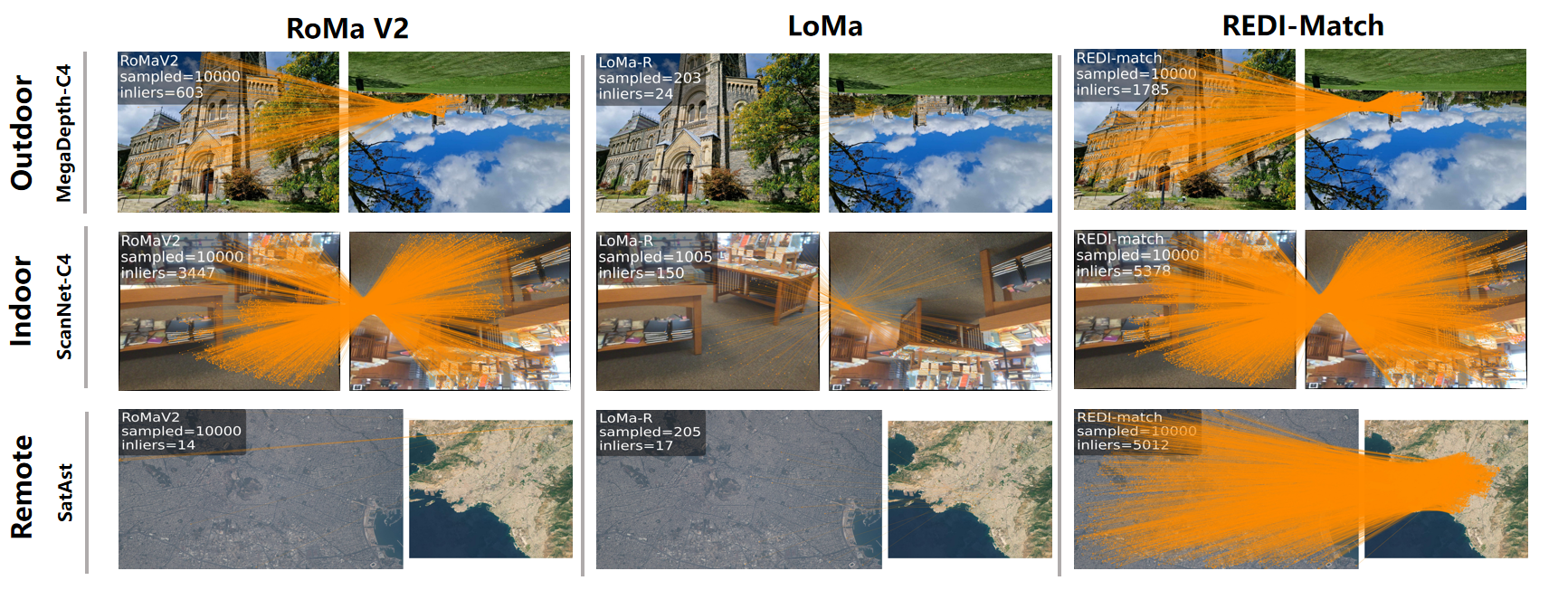}
    \caption{\textbf{ Qualitative comparison of REDI-Match against SOTA methods under severe in-plane rotations.} We evaluate across outdoor (MegaDepth-C4), indoor (ScanNet-C4), and remote sensing (SatAst) datasets. All input images are resized to $576 \times 576$. The reported inliers indicate the number of points after RANSAC, and orange lines visualize their one-to-one correspondences. For dense matchers (RoMa v2 and REDI-Match), visualizations are based on a fixed sampling of 10,000 points; for the sparse matcher (LoMa-R), the sample count reflects the detected keypoints. Notably, while existing SOTA methods suffer from catastrophic failure on the highly challenging SatAst dataset, REDI-Match consistently maintains dense and highly accurate alignments, demonstrating its robust zero-shot rotation generalization.}
    \label{fig:satast}
\end{figure*}

\section{Introduction}
\label{sec:intro}

Feature matching aims to establish reliable point correspondences across images, serving as a cornerstone for downstream tasks like Structure-from-Motion (SfM), 3D reconstruction, and robotic navigation~\cite{sarlinSuperGlueLearningFeature2020,sunLoFTRDetectorFreeLocal2021a}. Recently, dense matching has advanced this field by providing pixel-level correspondences~\cite{edstedt2023dkm,edstedtRoMaRobustDense2024}. This success is largely driven by Vision Foundation Models (VFMs)~\cite{caron2021emerging,DBLP:journals/tmlr/OquabDMVSKFHMEA24,simeoni2025dinov3}, which serve as powerful encoders capable of extracting deep semantics for high-accuracy matching.

However, severe in-plane rotation remains a critical bottleneck, degrading performance in applications such as aerial visual localization, remote sensing, and medical imaging~\cite{bokman2022case,lee2023learning}. To address this, recent research diverges into two paradigms. On one hand, data-driven methods rely on extensive data augmentation to implicitly learn rotational transformations~\cite{edstedt2025roma}. On the other hand, structure-driven methods employ Group Equivariant Convolutional Neural Networks (G-CNNs)~\cite{DBLP:conf/icml/CohenW16,DBLP:conf/iclr/CohenW17,DBLP:conf/nips/WeilerC19} to mathematically guarantee geometric consistency. Fundamentally, both approaches are trapped in a strict trade-off between semantic capacity and geometric consistency. Data-driven strategies require massive scaling of model parameters to accommodate rotational augmentations, yielding brittle generalization and computational inefficiency. Conversely, strictly equivariant networks struggle to match the rich semantic capacity of modern foundation models.

Driven by the necessity of these robust representations, modern frameworks typically freeze pre-trained VFMs. Consequently, state-of-the-art (SOTA) architectures resort to implicit data augmentation, shifting the entire burden of rotation generalization to the downstream decoder. To break this architectural bottleneck, we recognize that the geometric consistency of G-CNNs and the semantic capacity of VFMs are highly complementary. We propose a novel Rotation-Equivariant Distillation (REDI) paradigm. By distilling the representations of a heavy VFM into a lightweight, strictly rotation-equivariant encoder, we unify robust semantic representations with strict geometric consistency. This approach structurally models geometric variations early at the feature extraction stage, completely bypassing the need for massive data augmentation.

Building upon this paradigm, we introduce \textbf{REDI-Match}, an efficient dense matching framework tailored for rotation-robust applications. To exploit the equivariant features from our encoder, we equip the downstream decoder with a lightweight, \textbf{entropy-driven spatial alignment module}. Rather than forcing the network to implicitly learn geometric transformations, this module explicitly evaluates discrete rotation hypotheses. By utilizing information entropy to lock onto the canonical orientation, it eliminates global ambiguity before refining small-angle variations. Ultimately, this mathematically grounded pipeline achieves an optimal balance between computational efficiency and geometric robustness.

The main contributions of this paper are summarized as follows:
\begin{itemize}
    \item We propose \textbf{REDI}, an equivariant distillation paradigm that transfers the semantic knowledge of a VFM into a lightweight, strictly rotation-equivariant encoder, unifying robust semantics with strict geometric equivariance in a single feature manifold.
    \item We introduce an entropy-driven alignment module that resolves global rotation ambiguity by explicitly evaluating discrete hypotheses, eliminating the need for implicit data-fitting.
    \item We present \textbf{REDI-Match}, an efficient dense matching framework that delivers SOTA performance under severe rotations.Notably, it achieves a \textbf{13.89\%} absolute pose accuracy improvement on the highly challenging SatAst dataset while operating \textbf{1.9$\times$} faster than the current SOTA (RoMa v2), enabling real-time inference (\textbf{$\sim$41 FPS}) on a single RTX 4090 GPU.
\end{itemize}
\section{Related Work}
\label{sec:related}

\subsection{Feature Matching}
Feature matching methods can be broadly categorized into sparse, semi-dense, and dense approaches, which establish image correspondences at varying granularities ranging from discrete keypoints to global pixel-wise mappings. Early sparse methods~\cite{detone2018superpoint,rubleeORBEfficientAlternative2011,sarlinSuperGlueLearningFeature2020,lindenbergerLightGlueLocalFeature2023,chenRDDRobustFeature2025,liuLiftFeat3DGeometryAware2025a,pautratGlueStickRobustImage2023} rely on local associations for fast inference, albeit at the cost of lower precision. To bridge this gap, semi-dense methods~\cite{sunLoFTRDetectorFreeLocal2021a,tangQuadtreeAttentionVision2021,giang2023topicfm,wangEfficientLoFTRSemiDense2024,yuAdaptiveSpotGuidedTransformer2023,zhangLearningTwoViewCorrespondences2019,chenASpanFormerDetectorFreeImage2022a,zhouPatch2PixEpipolarGuidedPixelLevel2021a} emerged as a pragmatic middle ground, balancing matching performance with computational speed. Pushing toward ultimate accuracy, dense matching methods~\cite{edstedt2023dkm,truongGLUNetGlobalLocalUniversal2020} establish pixel-wise correspondences. Recently, frameworks like the RoMa series~\cite{edstedtRoMaRobustDense2024} have integrated rich semantic priors from VFMs. To further push performance boundaries, SOTA algorithms such as UFM~\cite{zhang2025ufm} and RoMa v2~\cite{edstedt2025roma} scale their training paradigms across extensively augmented, multi-domain datasets.

However, we observe that in pursuit of competitive accuracy, many sparse and semi-dense methods now adopt heavy modules that severely compromise their original real-time efficiency. Meanwhile, although dense matchers excel in fine-grained alignment, their massive parameter counts and heavy computational burdens, often exacerbated by reliance on VFMs, typically restrict them to offline applications. This unresolved trade-off between semantic capacity and real-time efficiency highlights the urgent need for a lightweight matching architecture capable of high-speed inference.

\subsection{Equivariant Networks and Rotation Consistency}
Rotation matching has long stood as a fundamental challenge in the field of feature matching. To overcome the rotational sensitivity of standard CNNs, G-CNNs~\cite{DBLP:conf/icml/CohenW16,DBLP:conf/iclr/CohenW17,DBLP:conf/nips/WeilerC19} mathematically guarantee geometric consistency. Efforts like Rotation-Steerers~\cite{bokman2024steerers} and co-design strategies~\cite{edstedt2024dedode,edstedt2024dedode3dv} attempt to inject these priors but achieve only approximate equivariance, failing under extreme baseline shifts. Alternatively, recent methods like LoMa-R~\cite{nordstrom2026who} attempt to improve robustness through extensive rotation augmentation across mixed datasets.

Despite these advancements, a fundamental dilemma persists. Strictly equivariant matchers require training from scratch, preventing them from leveraging the rich semantic capacity of modern VFMs. Conversely, post-hoc solutions relying on either approximate steering or heavy data augmentation inevitably degrade the geometric phase information critical for dense alignment. Consequently, they falter under the compound challenge of extreme rotations and large viewpoint changes. This exposes a critical structural gap: the inability to unify deep semantic power with rigorous geometric consistency within a single feature manifold.

\subsection{Vision Foundation Models and Knowledge Distillation}
VFMs such as DINO~\cite{caron2021emerging}, DINOv2~\cite{DBLP:journals/tmlr/OquabDMVSKFHMEA24}, and DINOv3~\cite{simeoni2025dinov3} have demonstrated remarkable semantic understanding, making them powerful backbones for dense prediction tasks. Downstream applications favor lightweight VFM variants compressed via Knowledge Distillation (KD)~\cite{hinton2015distilling}. Early KD methods focused on aligning intermediate feature representations~\cite{DBLP:journals/corr/RomeroBKCGB14}, while later works improved fidelity through similarity preservation~\cite{tung2019similarity} and relational consistency~\cite{park2019relational}.

Traditionally, KD has been treated exclusively as a utility for model compression, aimed at maximally retaining the teacher's performance. Breaking from this convention, we investigate a novel perspective: can KD act as a principled bridge to unify two fundamentally distinct feature manifolds? Specifically, we explore its potential to connect the non-equivariant semantic space of a teacher with the structured equivariant space of a student.

% Use figure* for multi-column figure

\begin{figure*}[t]
    \centering
    \includegraphics[width=\textwidth,height=0.28\textheight]{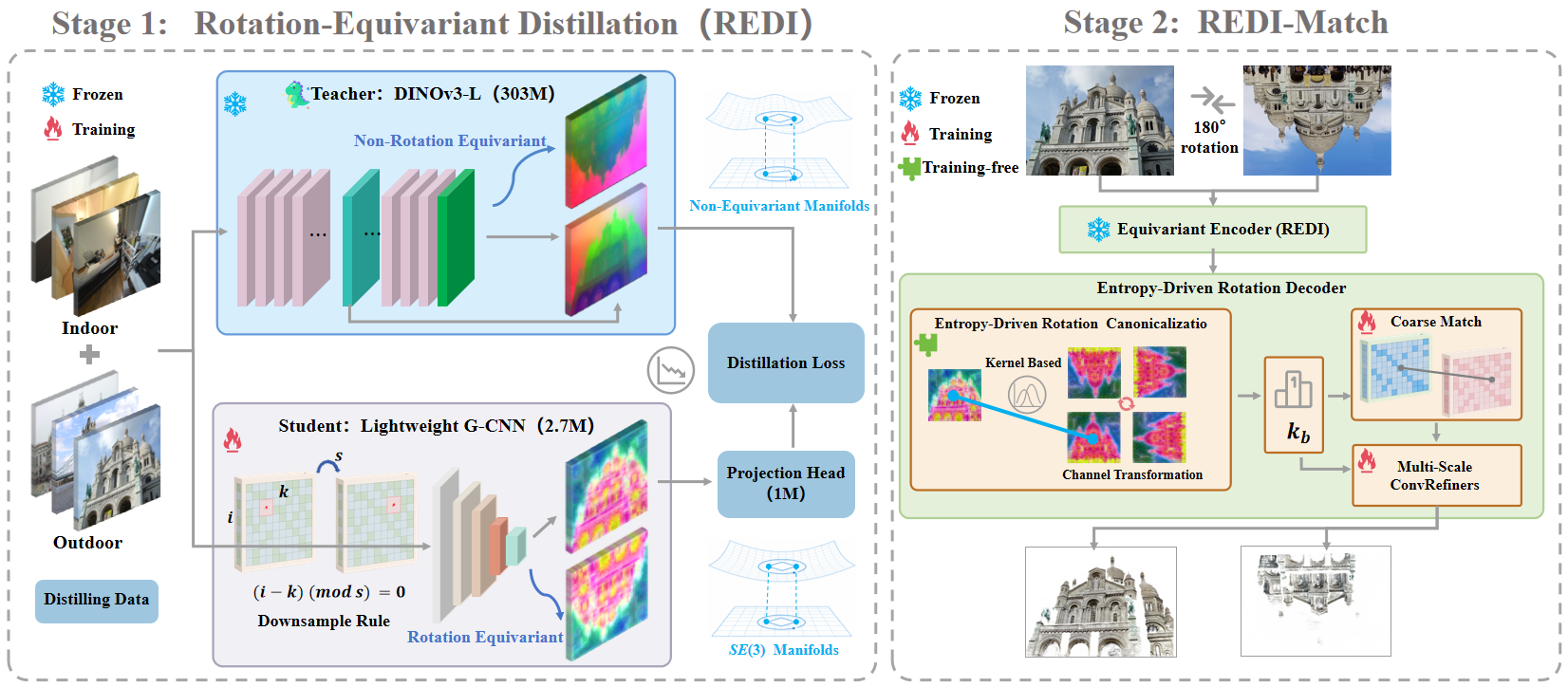}
    \caption{\textbf{Overview of the REDI-Match framework.} \textbf{(Left) Stage 1: Rotation-Equivariant Distillation (REDI).} A frozen VFM (DINOv3-L) is distilled into a lightweight G-CNN. Optimized via Mean Squared Error (MSE) loss to strictly preserve geometric phase information, this process projects non-equivariant semantics onto an $SE(3)$ manifold. \textbf{(Right) Stage 2: REDI-Match.} Given rotated image pairs, the frozen REDI encoder extracts equivariant features for the \textbf{Entropy-Driven Rotation Decoder}. Within this decoder, a training-free canonicalization module utilizes information entropy to evaluate discrete hypotheses, explicitly locking onto the canonical orientation ($k_b$). This eliminates global ambiguity early, allowing the trainable coarse matcher and multi-scale refiners to establish robust dense correspondences without rotation-augmented data.}
    \label{fig:pipeline}
\end{figure*}

\section{Method}
\label{sec:method}

\subsection{Overall Architecture}

To address severe in-plane rotations without relying on heuristic data augmentation, our framework builds upon an equivariant distillation architecture that bridges non-equivariant VFMs with G-CNNs. This design is fundamentally motivated by a key empirical observation: while standard dense matchers degrade catastrophically under extreme rotations (as evidenced in Tab.~\ref{tab:rotation_benchmark}), their native spatial correlation priors inherently tolerate continuous, small-angle perturbations . To achieve an optimal trade-off between robustness and computational efficiency, we abandon the prohibitive cost of continuous rotation group modeling. Instead, we propose a ``discrete canonicalization plus continuous residual refinement'' paradigm, strategically factoring out large-angle variations explicitly while leaving local deviations to be handled by the network's inherent capacity.

As illustrated in Fig.~\ref{fig:pipeline}, the matching pipeline begins with a lightweight, semantic-guided $C_N$-equivariant encoder. By distilling knowledge from a heavy VFM (Sec.~\ref{sec:distill}), this encoder structurally endows the features with explicit geometric priors, resolving arbitrary large-angle transformations into deterministic combinations of spatial rotations and group-channel permutations. At the coarsest feature scale, an entropy-driven module (Sec.~\ref{sec:rot_decoder}) evaluates $C_N$ discrete rotation hypotheses in parallel. Guided by the sharpness of the resulting matching distributions, it selects the optimal rotation index $k^*$, thereby performing global spatial canonicalization. This critical step eliminates the dominant large-scale orientation ambiguity and maps the features into a unified, gravity-aligned reference frame.

Operating entirely within this canonical space, the remaining relative rotation between the feature maps is strictly bounded within a small residual regime . In this constrained space, a multi-scale ConvRefiner executes top-down refinement to resolve local structural inconsistencies and continuous small-angle deviations. Benefiting from the initial discrete alignment, the refiner's native correlation mechanisms remain highly effective entirely without the need for rotation-based data augmentation during training. Finally, a decanonicalization step inversely transforms the refined dense correspondence field back to the original image coordinate system. As validated in Tab.~\ref{tab:rotation_benchmark}, this architecture yields unique advantages on continuous rotation benchmarks, consistently outperforming methods that rely solely on data-driven augmentation, under a decoupled two-stage optimization paradigm (Sec.~\ref{sec:two_stage}).

\subsection{Equivariant Distillation Framework}\label{sec:distill}

We distill semantic representations from the non-equivariant VFM DINOv3-L (teacher $\Phi_T$) into a lightweight G-CNN (student $\Phi_S$). To bridge the representational gap between the two, we introduce a standard linear projection head $P$. Since the teacher's feature space lacks strict rotational symmetry, this unconstrained linear projection provides the necessary flexibility to align the feature manifolds, avoiding the mapping capacity bottlenecks imposed by an equivariant projection head.

Although group convolutions exhibit algebraic equivariance, internal spatial subsampling operations (e.g., striding or pooling) introduce aliasing, which breaks grid-orbit consistency and degrades geometric phase information. To guarantee strict layer-wise equivariance under discrete transformation groups $G$ , we impose rigorous structural constraints on $\Phi_S$. Following architectural designs in standard equivariant networks~\cite{DBLP:conf/nips/WeilerC19}, for any subsampling layer with input dimension $i$ (including padding), kernel size $k$, and stride $s$, we enforce the following grid-alignment modulo condition:
\begin{equation}
(i - k) \bmod s = 0
\label{eq:grid_align}
\end{equation}
Crucially, while this constraint was initially proposed for standard supervised training ~\cite{DBLP:conf/iccvw/Edixhoven0G23}, we demonstrate its critical structural role within the REDI paradigm. By explicitly preventing spatial aliasing through this structural constraint, $\Phi_S$ satisfies the exact equivariance condition:
\begin{equation}
\Phi_S\bigl(\mathcal{T}_g x\bigr) = \mathcal{R}_g \Phi_S(x), \quad \forall g \in G
\label{eq:perfect_equivariance}
\end{equation}
where $\mathcal{T}_g$ denotes the spatial transformation applied to the input image, and $\mathcal{R}_g$ represents the corresponding group action in the feature space.

While $\Phi_S$ is architecturally equivariant, empirical results indicate that rotation-augmented distillation further improves robustness by explicitly aligning semantic manifolds across multiple orientations, rather than relying solely on architectural priors. Finally, to capture both fine-grained textures and global context, the student is aligned with a weighted aggregation of the teacher's multi-scale features. Let $\mathcal{L}$ denote the set of selected intermediate and final layers, and $w_l$ their corresponding weights. The aggregated multi-scale Teacher feature serves as the distillation target:
\begin{equation}
\Phi_T(x) = \sum_{l \in \mathcal{L}} w_l \Phi_T^{(l)}(x)
\label{eq:multiscale_teacher}
\end{equation}
A standard projection head $P$ maps the student output to this target space, thereby preserving the dense local saliency required for precise sub-pixel matching. The concrete optimization loss function is discussed in Sec.~\ref{sec:two_stage}.

\subsection{Entropy-Driven Rotation Decoder}\label{sec:rot_decoder}

We adopt Gaussian Process (GP) kernel regression for its inherent continuous spatial correlation priors, which stabilize predictions under large geometric deformations. However, existing kernel-based matchers lack explicit rotation awareness. To bridge this gap, we introduce our core contribution: an entropy-driven rotation estimation module.

Leveraging the equivariance property from Eq.~\ref{eq:perfect_equivariance}, we obtain feature maps at the coarsest scale corresponding to the four $C_4$ orientations by applying the group action $\mathcal{R}_k$ to the B-side features $F_B$:
\begin{equation}
Y^{(k)} = \mathcal{R}_k(F_B), \quad k \in \{0,1,2,3\},
\label{eq:rot_candidate}
\end{equation}
where $\mathcal{R}_k$ denotes the composite spatial rotation and group-channel permutation, and $k$ indexes $0^\circ,90^\circ,180^\circ,270^\circ$. For efficient global context, $F_A$ and $Y^{(k)}$ are uniformly downsampled via adaptive pooling to sequence tokens $X, Y^{(k)} \in \mathbb{R}^{B\times N\times d}$.

\begin{figure}[t]
    \centering
    \includegraphics[width=\columnwidth]{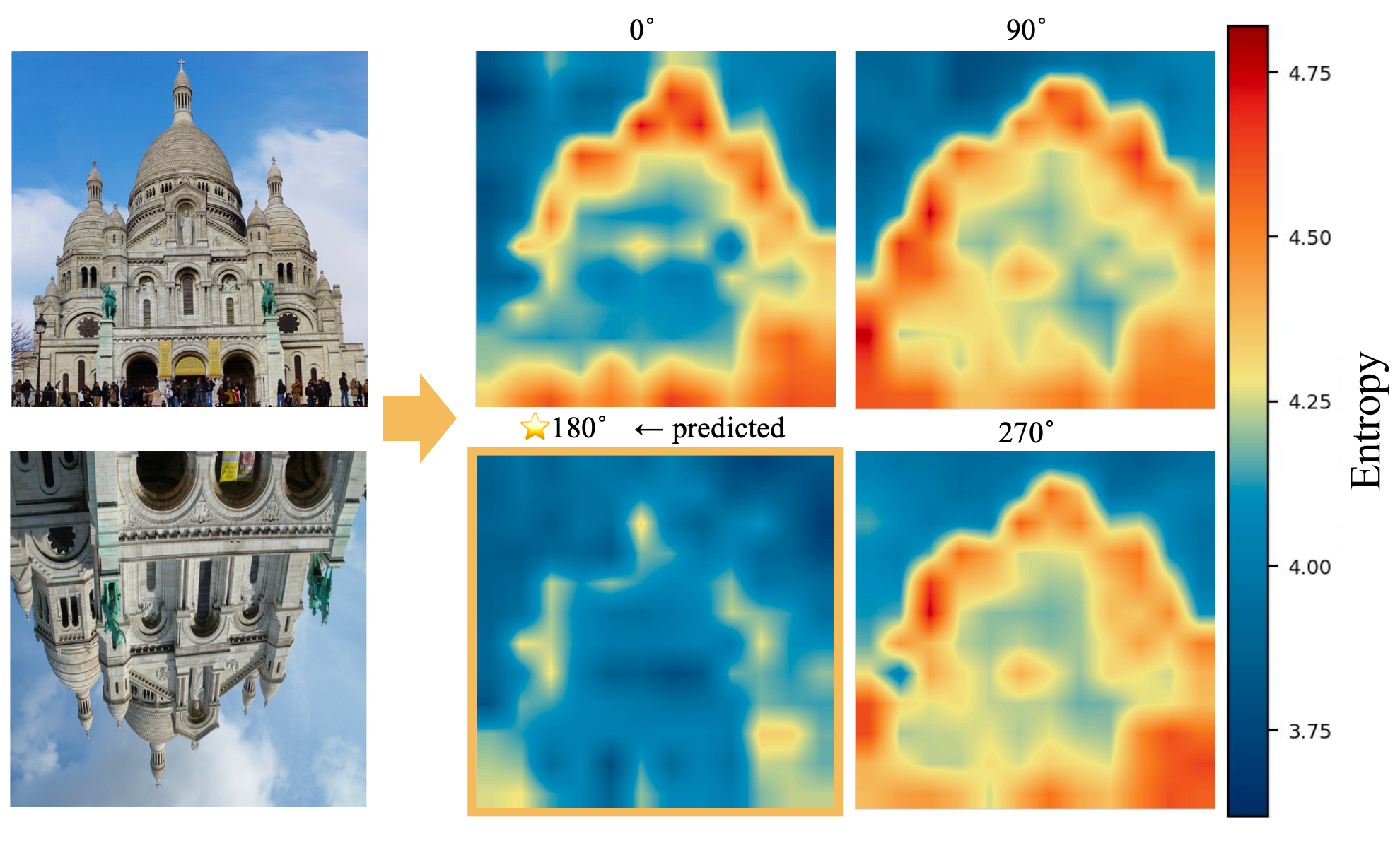}
    \caption{\textbf{ $C_4$ rotation detection via rotation kernel entropy discrimination.} Correct matching is consistently accompanied by a low-entropy distribution across orientation hypotheses. By evaluating the mean entropy $\bar{H}$, this module effectively identifies the optimal feature orientation, thereby resolving multi-modal distribution ambiguities in the dense matching process.}
    \label{fig:entropy_detection}
\end{figure}

As illustrated in Fig.~\ref{fig:entropy_detection}, the key insight is that the correct rotation yields a sharp, low-entropy matching distribution, while incorrect rotations produce diffuse, high-entropy affinities. We compute an exponential cosine affinity kernel with temperature $T>0$:
\begin{equation}
K_{b,ij}^{(k)} = \exp\left( \frac{\langle X_{b,i}, Y_{b,j}^{(k)} \rangle}{T \cdot \|X_{b,i}\|\,\|Y_{b,j}^{(k)}\|} \right),
\label{eq:kernel_matrix}
\end{equation}
and obtain the row-normalized matching probabilities
\begin{equation}
P_{b,ij}^{(k)} = \frac{K_{b,ij}^{(k)}}{\sum_{j'=1}^{M} K_{b,ij'}^{(k)}}.
\label{eq:prob_matrix}
\end{equation}
Matching certainty is quantified via mean Shannon entropy normalized by $\log M$. Let $\alpha = 1 / \log M$:
\begin{equation}
\tilde{H}_b^{(k)} = -\frac{\alpha}{N} \sum_{i,j} P_{b,ij}^{(k)} \log P_{b,ij}^{(k)}.
\label{eq:norm_entropy}
\end{equation}

We select the rotation minimizing entropy:
\begin{equation}
k_b = \argmin_{k\in\{0,1,2,3\}} \tilde{H}_b^{(k)}.
\label{eq:rot_argmin}
\end{equation}

% GP-based C4 rotation detection via entropy minimization

Once $k_b$ is identified, we apply the inverse transform $\mathcal{R}_{k_b}^{-1}$ to the multi-scale B-side features, projecting all branches into a unified gravity-aligned reference frame. Within this canonical space, the multi-scale ConvRefiner resolves residual local inconsistencies and small-angle deviations via top-down refinement, after which a decanonicalization step restores the dense flow and its certainty map to the original image coordinates.

Notably, this rotation estimation module is entirely training-free, requiring no parameter updates beyond the distilled geometric symmetry. The optimization objectives are discussed in Sec.~\ref{sec:two_stage}.

\subsection{Two-Stage Decoupled Optimization Paradigm}\label{sec:two_stage}

A defining characteristic of our framework is that rotational robustness is architecturally encoded rather than purely data-driven.

\noindent \textbf{Stage One: Equivariant Distillation Loss.} The distillation loss is a Mean Squared Error (MSE) objective applied on $C_4$-rotated views:
\begin{equation}
\mathcal{L}_{\mathrm{distill}} = \bigl\|P\bigl(\Phi_S(\mathcal{T}_g x)\bigr) - \Phi_T(\mathcal{T}_g x)\bigr\|_2^2.
\label{eq:distill_loss}
\end{equation}
Our ablation study (Sec.~\ref{sec:ablation}) identifies MSE as the most robust distillation objective among KL divergence, cosine similarity, and MSE. Unlike these alternatives, MSE jointly constrains both direction and magnitude of feature vectors, preserving the feature-magnitude saliency essential for sub-pixel alignment under extreme domain shifts. After distillation, $P$ is discarded and only $\Phi_S$ is retained.

\noindent \textbf{Stage Two: Dense Matching Loss.} The second stage optimizes the continuous correspondence estimator, comprising GP kernel regression and multi-scale ConvRefiner. The entropy-driven rotation decoder (Sec.~\ref{sec:rot_decoder}) is a purely training-free module---it solely exploits the geometric symmetry distilled in Stage One. Consequently, the matching decoder is trained exclusively under gravity-aligned states without any in-plane rotation augmentation. Following RoMa~\cite{edstedtRoMaRobustDense2024}, we employ a combined classification-regression loss to supervise dense flow predictions against ground-truth correspondences: the classification term optimizes match confidence, while the regression term minimizes the endpoint error of the dense warp.

This asymmetric decoupling yields a clear efficiency advantage: the Teacher is queried only once during distillation, and the matching decoder incurs zero rotation overhead at inference time, operating entirely under gravity-aligned canonical coordinates. 

\begin{table*}[t]
    \centering
    \caption{\textbf{Quantitative evaluation on discrete rotation datasets (MegaDepth-C4, ScanNet-C4, and HPatches-C4)} as well as continuous and extreme out-of-distribution rotation benchmarks (HPatches-Rot360 and SatAst). All tests are conducted at a resolution of $576 \times 576$. AUC scores are reported as raw values.}
    \label{tab:rotation_benchmark}
    \resizebox{\textwidth}{!}{%
        \begin{tabular}{l ccc ccc ccc ccc ccc}
            \toprule
            & \multicolumn{3}{c}{MegaDepth-C4 (AUC \% $\uparrow$)} & \multicolumn{3}{c}{ScanNet-C4 (AUC \% $\uparrow$)} & \multicolumn{3}{c}{HPatches-C4 (AUC \% $\uparrow$)} & \multicolumn{3}{c}{Roto360 (AUC \% $\uparrow$)} & \multicolumn{3}{c}{SATAST (AUC \% $\uparrow$)} \\
            \cmidrule(lr){2-4} \cmidrule(lr){5-7} \cmidrule(lr){8-10}  \cmidrule(lr){11-13} \cmidrule(lr){14-16}
            Method & @5$^\circ$ & @10$^\circ$ & @20$^\circ$ & @5$^\circ$ & @10$^\circ$ & @20$^\circ$ & @3$^\circ$ &  @5$^\circ$ &  @10$^\circ$  & @3$^\circ$ & @5$^\circ$ & @10$^\circ$ & @5$^\circ$ & @10$^\circ$ & @20$^\circ$\\
            \midrule
            \multicolumn{10}{l}{\textbf{Sparse}} \\
            \rowcolor{green!4} ALIKED+VGGT~\cite{zhao2023aliked,wang2025vggt}                     & 18.44 & 32.57 & 47.01 & 10.64 & 23.72 & 38.94 & 13.62 & 23.24 & 38.03 & 52.74 & 75.64 & 94.39 & 14.33 & 19.64 & 22.72 \\
            \rowcolor{green!4} RELF~\cite{lee2023learning}                          & 16.51 & 30.60 & 46.71 & 6.24 & 14.80 & 26.07 & 36.56 & 55.05 & 73.53 & 71.01 & 82.05 & 88.29 & 1.09 & 2.81 & 5.76 \\
            \rowcolor{green!4} Steerers$^\text{CVPR'24}$~\cite{bokman2024steerers}       & 48.72 & 65.88 & 78.92 & 14.70 & 29.60 & 44.47 & 46.17 & 63.37 & 79.03 & 89.90 & 92.06 & 94.38 & 1.92 & 2.90 & 3.70 \\
            \rowcolor{green!4} LoMa-R$^\text{CVPRW'26}$~\cite{nordstrom2026who}                & 45.20 & 62.61 & 76.63 & 17.13 & 31.94 & 45.38 & 34.32 & 49.76 & 65.46 & 45.24 & 52.01 & 61.15 & \underline{27.43} & \underline{41.98} & \underline{51.74} \\
            \midrule
            \multicolumn{10}{l}{\textbf{Semi-Dense}} \\
            \rowcolor{cyan!4} Se2-LoFTR$^\text{CVPRW'22}$~\cite{bokman2022case}     & 33.37 & 49.79 & 64.59 & 13.24 & 28.18 & 44.74 & 35.17 & 50.82 & 69.50 & 56.50 & 75.01 & 91.84 & OOD & 0.22 & 0.85 \\
            \midrule
            \multicolumn{10}{l}{\textbf{Dense}} \\
            \rowcolor{orange!10} DKM$^\text{CVPR'23}$~\cite{edstedt2023dkm}            & 16.20 & 20.57 & 23.77 & 12.92 & 24.52 & 35.70 & 23.30 & 28.93 & 34.83 & 48.40 & 50.15 & 52.04 & 9.89 & 12.12 & 13.57 \\
            \rowcolor{orange!10} RoMa$^\text{CVPR'24}$~\cite{edstedtRoMaRobustDense2024}           & 25.90 & 37.74 & 48.47 & 12.64 & 23.93 & 35.61 & 28.78 & 41.24 & 54.31 & 50.40 & 58.35 & 67.28 & 14.12 & 18.96 & 22.07 \\
            \rowcolor{orange!10} RoMa v2$^\text{arXiv'25}$~\cite{edstedt2025roma}        & \underline{53.45} & \underline{68.93} & \underline{80.09} & \underline{29.03} & \underline{49.47} & \underline{65.91} & \underline{68.17} & \underline{78.05} & \underline{87.01} & \underline{97.27} & \underline{97.65} & \underline{98.02} & 18.40 & 24.20 & 28.50 \\
            % UFM                             & 18.72 & 30.01 & 42.33 & 15.36 & 30.14 & 44.99 & 20.96 & 33.16 & 50.21 & - & - & - & - & - & - \\
            \rowcolor{red!18}
            REDI-Match                      & \textbf{59.22} & \textbf{74.27} & \textbf{84.76} & \textbf{29.48} & \textbf{50.60} & \textbf{67.98} & \textbf{70.41} & \textbf{79.59} & \textbf{87.71} & \textbf{98.45} & \textbf{98.57} & \textbf{98.64} & \textbf{41.32} & \textbf{50.62} & \textbf{57.37} \\
            \bottomrule
        \end{tabular}%
    }
\end{table*}

\section{Experiments}
\label{sec:experiments}
\subsection{Experimental Setup}

\noindent \textbf{Training Data.} The outdoor version is trained on MegaDepth~\cite{li2018megadepth} and the indoor version on a MegaDepth-ScanNet~\cite{dai2017scannet} mixture. As summarized in Tab.~\ref{tab:data_comparison}, REDI-Match requires only two standard datasets, in stark contrast to prior methods that rely on massive multi-dataset mixtures, underscoring that our robustness is method-driven rather than data-driven.

\begin{table}[t]
\centering
\caption{\textbf{Training data comparison.} REDI-Match achieves superior rotation robustness with only 2 datasets, versus 10--17 for prior matchers, demonstrating that our gains arise from equivariant distillation rather than data scaling.}
\label{tab:data_comparison}
\resizebox{\columnwidth}{!}{%
\footnotesize
\begin{tabular}{l p{5cm} c c}
\toprule
\textbf{Method} & \textbf{Training Datasets} & \textbf{Datasets} & \textbf{Paradigm} \\
\midrule
RoMa v2~\cite{edstedt2025roma}   & MegaDepth, ScanNet++ v2, AerialMD, BlendedMVS, Hypersim, \textit{+5 others} & 10 & Data-Driven \\
LoMa-R~\cite{nordstrom2026who} & MegaDepth, ScanNet++ v2, MegaScenes, Map-Free, TartanAir v2, \textit{+12 others} & 17 & Data-Driven \\
\midrule
\rowcolor{gray!15}
REDI-Match                      & \textbf{MegaDepth, ScanNet} & \textbf{2} & \textbf{Method-Driven} \\
\bottomrule
\end{tabular}%
}
\end{table}

\noindent \textbf{Evaluation Benchmarks.} We establish a multi-level benchmark~\cite{jin2021image,sattler2018benchmarking,taira2018inloc}: (1) standard and $C_4$ discrete rotation~\cite{balntas2017hpatches} matching on MegaDepth, ScanNet, and HPatches; (2) continuous full-angle rotation robustness on HPatches Rot360; and (3) extreme out-of-distribution generalization on SatAst~\cite{vuong2025aerialmegadepth}, where most existing architectures suffer  failure. To ensure fair comparison of inference speed and accuracy across model architectures, all methods are evaluated at a unified resolution of $576 \times 576$.

\noindent \textbf{Encoder Distillation.} A frozen DINOv3 ViT-L/16 ($\sim$303M) serves as the Teacher, with a lightweight $C_4$-equivariant VGG-FPN ($\sim$2.7M) as the Student. A projection head ($\sim$1.0M) bridges the two and is discarded after distillation, yielding $\sim112\times$ compression. Distillation uses only RGB images at $224 \times 224$ resolution, with MSE loss and continuous rotation augmentations, running for 100 epochs on three NVIDIA A100 GPUs (80GB) with DDP, bfloat16 mixed precision, the LARS optimizer, batch size 768, learning rate 4.0, and completes in approximately 1 day.

\noindent \textbf{Decoder Training.} The matching decoder is trained on four NVIDIA RTX 4090 GPUs at $560 \times 560$ resolution following the RoMa protocol with standard photometric augmentations, strictly excluding in-plane rotation augmentations, and completes in approximately 2 days with EMA.

\subsection{State-of-the-Art Comparison on Rotation Benchmarks}

As shown in Tab.~\ref{tab:rotation_benchmark}, existing matchers degrade significantly under $C_4$ discrete rotations on MegaDepth, ScanNet, and HPatches due to lacking explicit geometric constraints. In contrast, REDI-Match remains highly stable, achieving state-of-the-art performance across all discrete settings. Furthermore, our framework maintains consistent robustness under unconstrained continuous rotations (HPatches-Rot360). This confirms that the REDI-Match framework maintains excellent robustness against continuous rotations, despite not relying on massive data augmentation or complex continuous-equivariant modeling.

This advantage peaks on the challenging out-of-distribution SatAst dataset, where REDI-Match improves the AUC@$5^\circ$ from 27.43\% to 41.32\%. We attribute this exceptional zero-shot generalization to the decoder's compact two-stage logic. Entropy-driven canonicalization at the coarsest scale isolates large rotations, restricting the downstream ConvRefiner to a bounded residual range. Consequently, this structural prior enables superior geometric generalization without the massive data scaling required by data-driven methods like RoMa v2 or LoMa-R~\cite{edstedt2025roma,nordstrom2026who}.

\begin{table*}[t]
    \centering
    \caption{\textbf{Quantitative evaluation on MegaDepth, ScanNet, and HPatches datasets,} grouped into Sparse, Semi-Dense, and Dense categories, with additional model size and inference latency statistics. All tests are conducted at a resolution of $576 \times 576$. Latency is measured on a single NVIDIA RTX 4090 GPU.}
    \label{tab:accuracy_576}
    \resizebox{\textwidth}{!}{%
        \begin{tabular}{l ccc ccc ccc cc}
            \toprule
            & \multicolumn{3}{c}{MegaDepth (AUC \% $\uparrow$)} & \multicolumn{3}{c}{ScanNet (AUC \% $\uparrow$)} & \multicolumn{3}{c}{HPatches (AUC \% $\uparrow$)} & \multicolumn{2}{c}{Efficiency} \\
            \cmidrule(lr){2-4} \cmidrule(lr){5-7} \cmidrule(lr){8-10} \cmidrule(lr){11-12}
            Method & @5$^\circ$  & @10$^\circ$  & @20$^\circ$  & @5$^\circ$  & @10$^\circ$  & @20$^\circ$  & @3$^\circ$  &  @5$^\circ$  &  @10$^\circ$  & Params (M) $\downarrow$ & Latency (ms) $\downarrow$ \\
            \midrule
            \multicolumn{12}{l}{\textbf{Sparse}} \\
            \rowcolor{green!4} ALIKED$^\text{TIM'23}$+LG $^\text{CVPR'24}$~\cite{zhao2023aliked,lindenbergerLightGlueLocalFeature2023}       & 37.25 & 55.31 & 70.87 & 16.65 & 33.28 & 49.95 & 60.38 & 73.08 & 84.37 & 12.56 & 27.31 \\
            \rowcolor{green!4} DISK$^\text{NeurIPS'20}$+LG~\cite{tyszkiewiczDISKLearningLocal2020,lindenbergerLightGlueLocalFeature2023}         & 36.38 & 54.64 & 70.32 & 14.37 & 28.90 & 44.29 & 49.66 & 64.43 & 79.38 & 12.98 & 31.79 \\
            \rowcolor{green!4} SP$^\text{CVPRW'18}$+LSD+GT~\cite{detone2018superpoint}       & 42.75 & 59.99 & 74.20 & 16.34 & 34.46 & 52.71 & 42.25 & 59.15 & 76.52 & 18.24 & 242.52 \\
            \rowcolor{green!4} ALIKED+VGGSFM $^\text{CVPR'24}$~\cite{zhao2023aliked,wang2025vggt}    & 37.05 & 55.05 & 69.80 & 10.89 & 21.71 & 32.73 & 51.38 & 66.25 & 80.07 & 51.41 & 780.27 \\
            \rowcolor{green!4} Steerers$^\text{CVPR'24}$~\cite{bokman2024steerers}        & 50.64 & 67.44 & 80.24 & 14.90 & 29.60 & 44.60 & 67.70 & 77.47 & 86.80 & 33.44 & 159.38\\
            \rowcolor{green!4} LoMa-R$^\text{CVPRW'26}$~\cite{nordstrom2026who}            & 50.06 & 66.82 & 79.81 & 25.93 & 47.70 & 66.55 &61.13 & 73.79 & 85.38 &341.52 &91.94\\
            \rowcolor{green!4} LoMa-G$^\text{arXiv'26}$~\cite{nordström2026lomalocalfeaturematching}            & 51.38 & 67.76 & 80.50 & 28.13 & 50.52 & 68.97 &59.32 & 72.79 & 84.82 &518.89 &141.63\\
            \midrule
            \multicolumn{12}{l}{\textbf{Semi-Dense}} \\
            \rowcolor{cyan!4} CasMTR$^\text{CVPR'24}$~\cite{caoImprovingTransformerbasedImage2023}          & 52.66 & 69.02 & 81.21 & 26.15 & 46.24 & 63.61 & 67.77 & 77.61 & 86.59 & 14.70 & 55.93 \\
            \rowcolor{cyan!4} CoMatch$^\text{ICCV'25}$~\cite{liCoMatchDynamicCovisibilityAware2025}         & 50.89 & 67.57 & 79.87 & 20.96 & 38.89 & 55.50 & 68.12 & 77.52 & 86.43 & \underline{12.01} & 29.58 \\
            \rowcolor{cyan!4} EfficientLoFTR$^\text{CVPR'24}$~\cite{wangEfficientLoFTRSemiDense2024}  & 48.48 & 65.40 & 78.31 & 18.66 & 36.57 & 53.55 & 68.26 & 77.57 & 86.17 & 15.05 & \textbf{18.22} \\
            \rowcolor{cyan!4} ASpanFormer$^\text{ECCV'22}$~\cite{chenASpanFormerDetectorFreeImage2022a}     & 56.41 & 72.56 & 83.91 & 26.26 & 46.69 & 64.07 & 68.16 & 77.47 & 86.11 & 15.76 & 93.10 \\
            \rowcolor{cyan!4} PMatch$^\text{CVPR'23}$~\cite{zhuPMatchPairedMasked2023}          & 55.53 & 71.41 & 82.95 & 27.04 & 47.83 & 65.38 & 39.48 & 43.22 & 46.62 & 15.75 & 37.46 \\
            \rowcolor{cyan!4} Se2-LoFTR$^\text{CVPRW'22}$~\cite{bokman2022case}       & 47.88 & 64.60 & 77.58 & 17.77 & 35.13 & 51.97 & 67.11 & 76.76 & 85.96 & \textbf{9.78} & 76.04 \\
            \midrule
            \multicolumn{12}{l}{\textbf{Dense}} \\
            \rowcolor{orange!10} DKM$^\text{CVPR'23}$~\cite{edstedt2023dkm}             & 58.13 & 73.34 & 84.11 & 29.22 & 50.36 & 68.00 & \underline{71.33} & \underline{80.40} & 88.42 & 72.26 & 34.25 \\
            \rowcolor{orange!10} RoMa$^\text{CVPR'24}$~\cite{edstedtRoMaRobustDense2024}            & 59.56 & 74.42 & 84.85 & \underline{30.99} & \underline{52.58} & \underline{70.19} & 70.35 & 79.88 & 88.20 & 415.66 & 78.40 \\
            \rowcolor{orange!10} RoMa v2$^\text{arXiv'25}$~\cite{edstedt2025roma}          & \textbf{60.02} & \textbf{74.77} & \textbf{85.19} & \textbf{33.27} & \textbf{55.96} & \textbf{73.40} & 70.74 & 80.36 & \underline{89.03} & 425.42 & 45.73 \\
            \rowcolor{orange!10} UFM $^\text{NeurIPS'25}$~\cite{zhang2025ufm}          & 34.08 & 51.00 & 66.45 & 28.11 & 50.65 & 69.12 & 58.60 & 70.44 & 82.40 & 477.76 & 95.18 \\
            \rowcolor{red!18}
            REDI-Match      & \underline{59.75} & \underline{74.70} & \underline{84.89} & 30.21 & 51.41 & 68.79 & \textbf{72.08} & \textbf{81.17} & \textbf{89.06} & 85.40 & \underline{24.05} \\
            \bottomrule
        \end{tabular}%
    }
\end{table*}

\subsection{Generalization on Standard Benchmarks and Efficiency}

On standard benchmarks (Tab.~\ref{tab:accuracy_576}), our framework delivers leading performance across all three datasets. On HPatches, it ranks first with 72.08/81.17/89.06 AUC at $3^\circ$/$5^\circ$/$10^\circ$; on MegaDepth, it trails RoMa v2 by less than 0.5\% ,and on ScanNet it achieves competitive performance,all while using $5\times$ fewer parameters (85M vs.\ 425M). This confirms that embedding VFM semantics into an equivariant manifold incurs no meaningful degradation on gravity-aligned scenes.

In terms of efficiency, the full pipeline achieves 24.05 ms (41 FPS) inference latency on a single NVIDIA RTX 4090, making it the fastest dense feature matcher to date. Combined with a compact 85.40M parameter footprint, the framework is well-suited for real-time deployment on consumer-grade and edge hardware.

\subsection{Equivariance Analysis }  
% Use figure* for multi-column figure
\begin{figure*}[t]
    \centering
    \includegraphics[width=\textwidth,height=0.26\textheight,keepaspectratio]{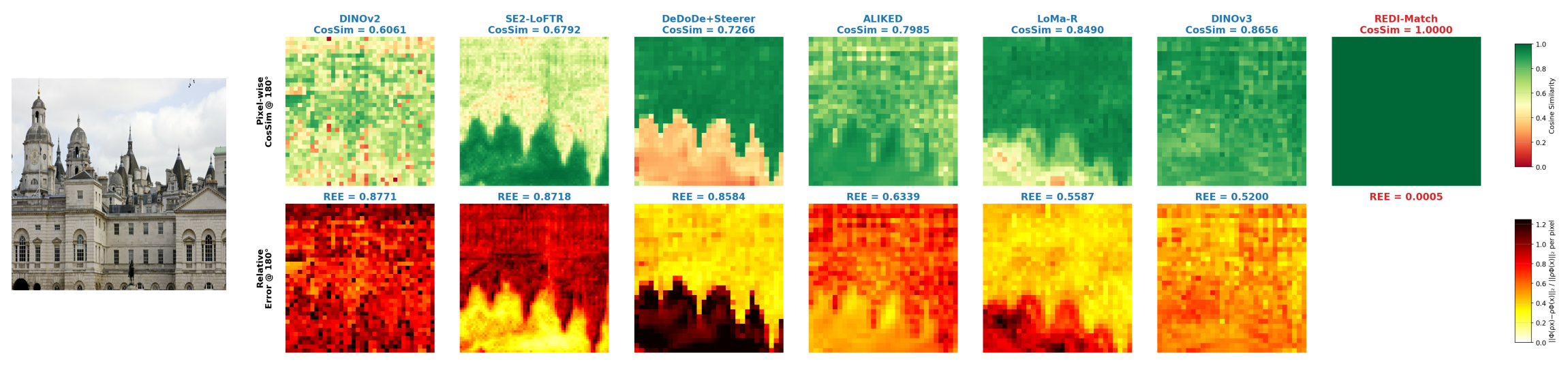}
    \caption{\textbf{Pixel-wise REE and PCS visualization of mainstream rotation matching encoders.}  We evaluate the rotation equivariance error and patch cosine similarity of current mainstream encoders under rotation. Our distilled equivariant encoder consistently preserves structural symmetry, achieving near-zero feature degradation. This is crucial for achieving efficient and robust rotation matching in subsequent stages.}
    \label{fig:pixelwise_REE-cosine_maps}
\end{figure*}
 
We quantitatively validate the strict equivariance properties of our encoder, maintaining machine-precision zero error and perfect spatial alignment across all discrete angles. As visualized in Fig.~\ref{fig:pixelwise_REE-cosine_maps}, we compare the feature distributions of different models under rotation. Standard non-equivariant foundation models such as DINOv3 exhibit severe feature distortions and structural collapse when subjected to in-plane rotations, resulting in high Pixel-wise Relative Equivariance Error (REE) and degraded Pixel-wise Cosine Similarity (PCS). In stark contrast, our equivariant encoder maintains strict spatial and semantic consistency across varying angles, visually confirming its immunity to geometric perturbations.

\begin{table}[t]
    \centering
    \caption{\textbf{Ablation study.} Evaluated on SatAst and MegaDepth-C4. \textit{Grad. Expl.}: gradient explosion.}
    \label{tab:ablation_master}
    \resizebox{\columnwidth}{!}{\footnotesize
        \begin{tabular}{l cccc}
            \toprule
            \multirow{2}{*}{\textbf{Variant}} & \multicolumn{2}{c}{\textbf{SatAst (AUC \% $\uparrow$)}} & \multicolumn{2}{c}{\textbf{Mega-C4 (AUC \% $\uparrow$)}} \\
            \cmidrule(lr){2-3} \cmidrule(lr){4-5}
            & @5$^\circ$ & @10$^\circ$ & @5$^\circ$ & @10$^\circ$ \\
            \midrule
            \multicolumn{5}{l}{\textbf{Encoder Distillation}} \\
            \midrule
            \abltarget{1}~~\textit{Model Design} \\
            \quad \quad w/o $(i-k) \bmod s = 0$  & 14.33 & 20.68 & 21.44 & 36.07 \\
            \midrule
            \abltarget{2}~~\textit{Training Strategy} \\
            \quad \quad w/o Distill       & \multicolumn{2}{c}{\textit{Grad. Expl.}} & \multicolumn{2}{c}{\textit{Grad. Expl.}} \\
            \midrule
            \abltarget{3}~~\textit{Loss Function} \\
            \quad \quad MSE $\rightarrow$ KL        & 17.59 & 21.12 & 57.74 & 72.72 \\
            \quad \quad MSE $\rightarrow$ Cosine     & 13.67 & 17.33 & 57.91 & 72.93 \\
            \midrule
            \abltarget{4}~~\textit{Projection Head} \\
            \quad \quad MLP $\rightarrow$ Equiv. Head       & 15.73 & 20.81 & 56.30 & 71.46 \\
            \midrule
            \abltarget{5}~~\textit{Data Strategy} \\
            \quad \quad Rot $\rightarrow$ Upright  & \underline{40.30} & \underline{50.00} & \underline{58.91} & \underline{74.00} \\
            \midrule
            \abltarget{6}~~\textit{Teacher Model} \\
            \quad \quad DINOv3-L $\rightarrow$ DINOv2-L  & 38.77 & 48.52 & 57.49 & 72.90 \\
            \midrule
            \multicolumn{5}{l}{\textbf{Decoder Training}} \\
            \midrule
            \abltarget{7}~~\textit{Rotation Decoder} \\
            \quad \quad w/o Entropy Rot. module & 16.42 & 19.48 & 16.94 & 21.52  \\
            \midrule
            \abltarget{8}~~\textit{Matching Backend} \\
            \quad \quad GP $\rightarrow$ Cross-Attn.   & 28.22 & 32.50 & 55.60 & 71.30 \\
            \midrule
            \rowcolor{gray!10}
            \textbf{REDI-Match (Full )} & \textbf{41.32} & \textbf{50.62}   & \textbf{59.22} & \textbf{74.27} \\
            \bottomrule
        \end{tabular}
    }
\end{table}

\subsection{Ablation Studies}\label{sec:ablation}
We conduct ablations on SatAst and MegaDepth-C4, categorized into encoder distillation and decoder design (Tab.~\ref{tab:ablation_master}).

\noindent \textbf{Encoder Distillation.} Violating the grid-alignment modulo condition (Eq.~\ref{eq:grid_align}) \abllink{1} reduces the encoder from strictly equivariant to only approximately equivariant, significantly degrading downstream matching accuracy. Training from scratch without distillation \abllink{2} leads to gradient explosion, indicating that the REDI paradigm provides necessary optimization stability. Regarding the loss function \abllink{3}, distribution-matching objectives like KL divergence and cosine similarity achieve performance comparable to MSE on the seen domain (MegaDepth-C4), but suffer from substantial degradation on the unseen remote sensing dataset (SatAst). This discrepancy indicates that MSE facilitates superior zero-shot generalization across domain gaps. By operating in the ambient Euclidean space and explicitly preserving geometric phase information, MSE prevents the student model from overfitting to the source domain's activation statistics, thereby retaining robust cross-domain transferability.

Furthermore, replacing the standard MLP projection head with an equivariant projection head \abllink{4} decreases performance. This indicates that a flexible, unconstrained MLP is necessary to bridge the structural gap between the student's equivariant features and the teacher's non-equivariant target space. Enforcing strict equivariance at the projection stage severely bottlenecks the distillation process, as it inherently struggles to map to the teacher's unconstrained representations. For data augmentation, rotation-augmented distillation \abllink{5} improves out-of-domain robustness compared to upright-only training. Finally, utilizing DINOv2-L instead of DINOv3-L as the teacher \abllink{6} causes a modest decline, demonstrating that the framework generalizes across different VFMs while still benefiting from stronger feature representations.

\noindent \textbf{Decoder Design.} Integrating the entropy-driven spatial alignment module into the decoder forms a compact two-stage logic. Removing this module \abllink{7} causes the most significant performance degradation, indicating the necessity of explicit spatial canonicalization under extreme rotations. Additionally, substituting GP kernel regression with standard implicit cross-attention \abllink{8} not only doubles training memory but also notably reduces accuracy. This implies that kernel-based spatial correlation priors provide more effective regularization for rotation-robust generalization than unconstrained attention mechanisms.
\section{Conclusion}
\label{sec:conclusion}

We present REDI-Match, an efficient dense matching framework that achieves robust in-plane rotation invariance through architectural equivariance rather than data augmentation. At its core, our REDI paradigm utilizes MSE to embed the rich semantics of VFMs into a lightweight G-CNN while preserving geometric phase information for robust cross-domain generalization. To exploit these equivariant features efficiently, we embed a training-free, entropy-driven alignment module directly into the decoder, establishing a compact two-stage logic of discrete global canonicalization followed by continuous residual refinement that seamlessly resolves rotational ambiguities.

Extensive evaluations confirm that REDI-Match establishes new SOTA performance on rotation benchmarks. Most notably, it yields a 13.89\% absolute accuracy improvement on the highly challenging SatAst dataset while achieving real-time inference at 41 FPS (24.05 ms). 

\noindent \textbf{Limitations and Future Work.} Handling extreme low-texture environments and full 3D viewpoints remains challenging. Future research will focus on expanding these planar constraints to full six-degree-of-freedom spatial symmetries, potentially via self-supervised depth integration, to further advance robust geometric matching.

\section*{Acknowledgements}
This work was supported by the National Key Research and Development Program of China (No. 2023YFB3906300), Tsinghua University Initiative Scientific Research Program, the New Cornerstone Science Foundation through the XPLORER PRIZE, and the Fundamental and Interdisciplinary Disciplines Breakthrough Plan of the Ministry of Education of China under Grant JYB2025XDXM109.

{\small
\bibliographystyle{ieeenat_fullname}
\bibliography{11_references}
}

\end{document}